\documentclass[sigconf]{acmart}
\usepackage{xcolor}
\usepackage{listings}
\usepackage[most]{tcolorbox}
\usepackage{hyperref}

\lstdefinestyle{py}{
  language=Python,
  basicstyle=\ttfamily\small\linespread{0.95}\selectfont,
  columns=fullflexible,
  keepspaces=true,
  basewidth=0.48em,
  numbers=none,
  lineskip=0pt,
  breaklines=true,
  showstringspaces=false,
  keywordstyle=\color{blue!70!black},
  commentstyle=\color{green!50!black},
  stringstyle=\color{cyan!45!black},
  numberstyle=\color{orange!70!black},
  identifierstyle=\color{black},
  emphstyle=\color{purple!70!black},
  emph={self,True,False,None},
  tabsize=2
}

\newtcblisting{pycode}[1][]{
  title=Python,
  colback=blue!2,
  colframe=blue!30!black,
  listing only,
  listing options={style=py},
  breakable,
  #1
}

\AtBeginDocument{%
  }

\setcopyright{acmlicensed}
\copyrightyear{2018}
\acmYear{2026}
\acmDOI{}
\acmConference[SIGSPATIAL '26]{34th ACM International Conference on
Advances in Geographic Information Systems}{Nov 03--06,
  2026}{Riverside, CA}
\acmISBN{978-1-4503-XXXX-X/2018/06}

\begin{document}

\title{Any Model, Any Place, Any Time: Get Remote Sensing Foundation Model Embeddings On Demand}

\author{Dingqi Ye, Daniel Kiv, Wei Hu, Jimeng Shi, Shaowen Wang}

\affiliation{%
  \institution{University of Illinois Urbana-Champaign}
  \city{Urbana}
  \state{Illinois}
  \country{USA}
}

\email{{dingqi2,dkiv2,weih9,jimeng8,shaowen}@illinois.edu}

\renewcommand{\shortauthors}{Dingqi et al.}

\begin{abstract}
 The remote sensing community is witnessing a rapid growth of foundation models, which provide powerful embeddings for a wide range of downstream tasks. However, practical adoption and fair comparison remain challenging due to substantial heterogeneity in model release formats, platforms and interfaces, and input data specifications. These inconsistencies significantly increase the cost of obtaining, using, and benchmarking embeddings across models. To address this issue, we propose {\itshape rs-embed}, a Python library that offers a unified, region of interst (ROI) centric interface: with a single line of code, users can retrieve embeddings from any supported model for any location and any time range. The library also provides efficient batch processing to enable large-scale embedding generation and evaluation. The code is available at: \textcolor{blue}{\textit{\url{https://github.com/cybergis/rs-embed}}}
\end{abstract}


\begin{teaserfigure}
  \centering
  \includegraphics[width=0.95\textwidth]{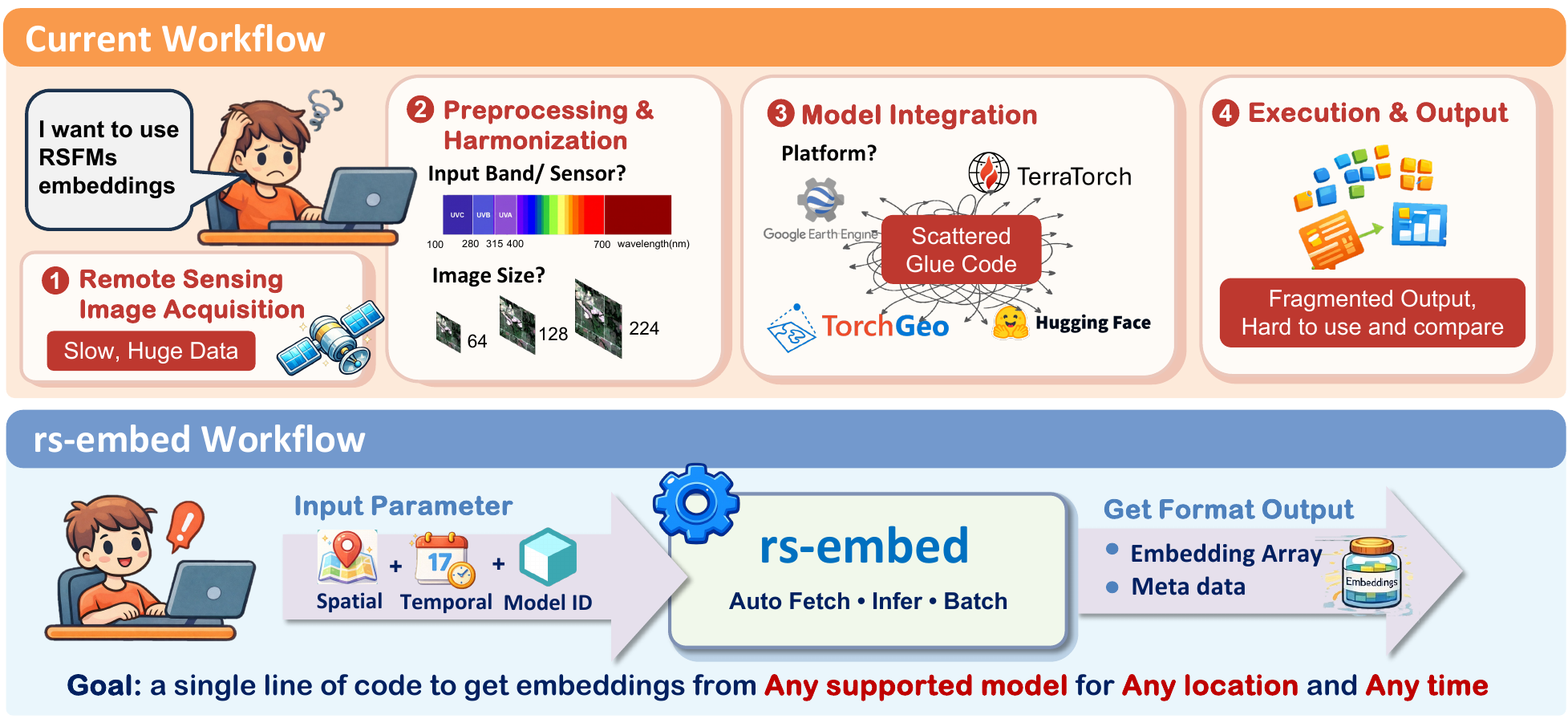}
  \caption{Comparison of a complex current workflow vs. rs-embed, which generates standardized embeddings (plus metadata) from location, time, and model ID in a single line of code.}
  \label{fig:overall}
\end{teaserfigure}


\maketitle

\section{Introduction}

Recent years have witnessed the rapid rise of remote sensing foundation models (RSFMs). By leveraging large-scale multimodal datasets, large-capacity models, and self-supervised pretraining, RSFMs acquire robust and generalizable representations, enabling strong transferability and wide applicability across diverse downstream tasks.

However, RSFMs still face practical challenges in accessibility, reusability, and comparability. First, release practices vary widely: some works only provide precomputed embeddings\cite{brown2025alphaearth}, while others only release the model\cite{szwarcman2025prithvi,liu2024remoteclip,cong2022satmae}, which users have to fetch the imagery and run inference themselves. Second, distribution and deployment remain fragmented: some models adopt standardized interfaces (e.g., Hugging Face), while others rely on custom repositories or specific framework versions—raising configuration and compatibility costs. Third, inconsistent input definitions and preprocessing (e.g., RGB\cite{cong2022satmae}, 6-band\cite{szwarcman2025prithvi} or 12-band\cite{jakubik2025terramind} Sentinel-2, MODIS\cite{spradlin2024satvision}) directly affect architectural adaptation, complicating fair downstream comparisons. Collectively, these issues increase the application difficulty and hinder unified evaluation and benchmarking.

\begin{figure}[h]
  \centering
  \includegraphics[width=0.47\textwidth]{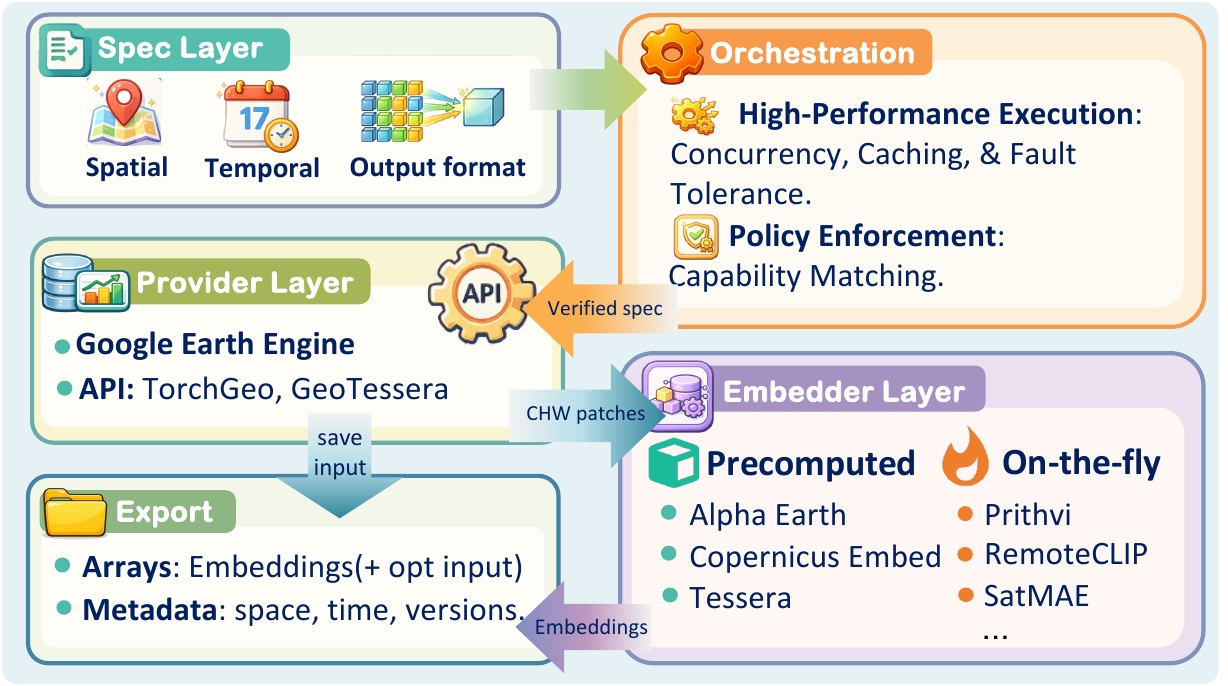}
  \caption{rs-embed architecture overview.}
  \label{fig:arc}
  
\end{figure}

To address these issues, we introduce {\itshape rs-embed} (Fig. \ref{fig:arc}), a Python library that centers the workflow on the user's region of interest (ROI). With a single line of code, users can uniformly obtain embeddings from diverse remote-sensing foundation models for any location and time range, substantially reducing invocation and configuration overhead. The library also provides efficient batching and engineering optimizations for large-scale processing, improving embedding throughput and scalability. Overall, rs-embed offers a unified and convenient toolkit for using, testing, and benchmarking remote-sensing foundation models.

\section{Method}



\subsection{Specification}


\textbf{Spatial Spec}:
Spatial extent can be specified by bounding boxes (\texttt{BBoxes}) or point buffers (\texttt{PointBuffer}). The spec includes a CRS (e.g., EPSG:4326) and geometry parameters, which are validated pre-execution (lat/lon ranges and ordering, positive buffer radius, and limits on area/scale).

\textbf{Temporal Spec}:
Time extent can be defined by year (\texttt{Temporal
-Spec.year}) or time ranges (\texttt{TemporalSpec.range}). It defines a left-closed, right-open interval $[start, end)$, rejecting invalid dates and reversed ranges. Together with the observation synthesis strategy (e.g., median/mosaic after time-window filtering), it determines observation construction logic and ensures reproducibility under the same spec.

\textbf{Output Spec}:
The Output Spec defines the embedding’s shape and aggregation. Pooled mode aggregates spatial dimensions into a fixed-length vector $z \in \mathbb{R}^d$ (e.g., mean pooling) for retrieval or tabular tasks. Grid mode outputs $z \in \mathbb{R}^{h \times w \times d}$ to preserve spatial context for pixel/grid-level models. Except for AlphaEarth and Tessera, whose embeddings are trained to be resolution-aligned with the input, $h \times w$ is generally not aligned with the input resolution $H \times W$ (Visualization on \ref{sec:visual}). It is introduced to represent features in a 3D tensor form that facilitates downstream convolutional processing.

\textbf{Sensor Spec}:
The Sensor Spec defines the raw imagery required for on-the-fly inference, including the data source (collection), bands, resolution (\texttt{scale\_m}), cloud limit (\texttt{cloudy\_pct}), \texttt{fill\_value}, and compositing method (median/mosaic). It specifies the dataset selection and preprocessing/compositing logic for the backend (e.g., GEE), and optionally enables pre-inference input quality checks via \texttt{check\_input}.

\subsection{Provider Layer}

The Provider Layer decouples heterogeneous remote-sensing data sources from model inference by offering a unified access interface that wraps cloud APIs (e.g., Google Earth Engine) into standardized numeric tensors. It handles projection and resampling, sensor-spec–guided spatiotemporal filtering and compositing (median or mosaic) to produce input patches, and converts observations into a consistent $(C, H, W)$ NumPy/Xarray format. This modular design also hides authentication/query complexity and enables straightforward extension to other platforms such as Microsoft Planetary Computer.

\subsection{Embedder Layer}
The Embedder Layer is the core engine for geospatial feature extraction. It uses an object-oriented design with a standardized foundation-model base class to uniformly wrap heterogeneous remote-sensing models.

\textbf{Base Model interface}:
The Embedder Layer defines a unified Embedder base class with standard APIs (e.g., \texttt{get\_embedding}, \texttt{get\_embeddings\_batch}, \texttt{describe}). Each foundation model implements these methods to expose a consistent interface while encapsulating model-specific details such as feature extraction, scale alignment, and band mapping.

\textbf{On-the-fly vs. Precomputed models}:{\itshape rs-embed} handles two acquisition modes. On-the-fly models run forward inference on raw imagery provided by the Provider Layer, applying preprocessing such as normalization/augmentation and optionally caching input patches for traceability. Precomputed models query embeddings already stored in the cloud (e.g., Alpha Earth), using SpatialSpec to locate, fetch, and assemble the required embedding vectors without executing a deep learning computation graph.

\subsection{Orchestration}

\begin{figure}
  \centering
  \includegraphics[width=0.48\textwidth]{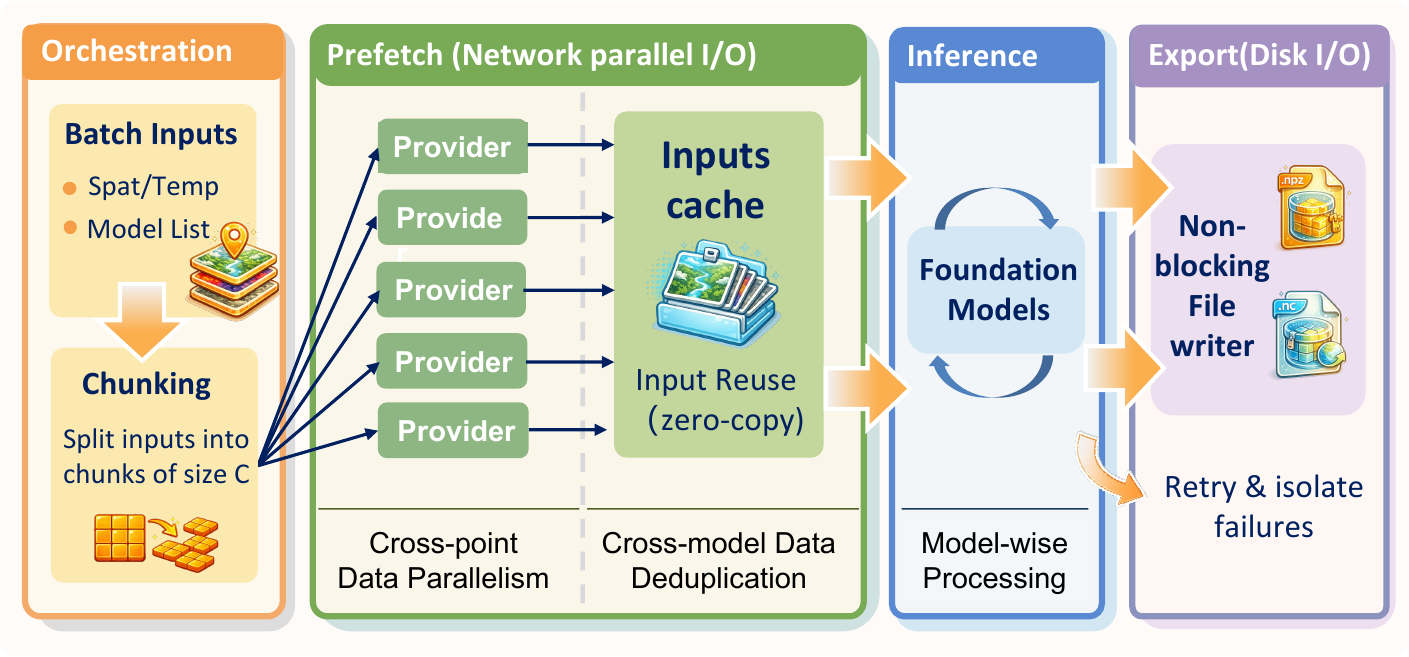}
  \caption{Parallel batch export pipeline (orchestration + I/O prefetch + inference + async export) that decouples network, compute, and disk writes to sustain high throughput reliably.}
  \label{fig:pipeline}
  
\end{figure}

\subsubsection{High-Performance Execution}

The {\itshape rs-embed} research has implemented a highly optimized parallel processing pipeline (as shown in Figure \ref{fig:pipeline}), consisting of 4 stages: Orchestration, Prefetch, Inference, Export. The goal is to maximize the utilization of the network and computing resources while ensuring the consistency of the results, and to reduce the costs caused by redundant downloads and repeated initializations.

\textbf{Orchestration}: The system ingests batched spatial queries (spatials), model sets (models), temporal constraints (temporal), and output specifications (output), and validates parameters. It then splits the workload into sub-batches by \texttt{chunk\_size} to limit peak memory and improve long-run stability. For each model, it resolves the effective sensor configuration (global or per-model override), enabling consistent indexing for cross-model input reuse.

\textbf{Prefetch}: In Provider(e.g. GEE), inputs are keyed by (point, sensor) and deduplicated per chunk to avoid redundant downloads. Unique requests are fetched in parallel via a thread pool (\texttt{num\_workers}), and the prefetched data are cached for reuse in persistence and inference, reducing end-to-end I/O overhead.

\textbf{Inference}:
Inference reuses cached embedder instances to avoid repeated weight loading and runtime initialization. For models supporting batch APIs, it prioritizes \texttt{get\_embeddings\_batch} to improve throughput and reduce Python overhead, with a fallback to per-sample inference for robustness. On-the-fly models directly consume prefetched inputs to avoid additional network reads.

\textbf{Export}: The system exports to npz/netcdf and overlaps I\/O with computation via asynchronous writing (\texttt{async\_write}, \texttt{writer\_workers}) in \texttt{out\_dir} mode for higher throughput.

\subsubsection{Policy enforcement \& Capability matching}

To ensure consistent and reproducible remote-sensing foundation model invocation, {\itshape rs-embed} uses a two-stage control flow: 
Policy Enforcement and Capability Matching. Policy enforcement pre-validates spatial, temporal, and output settings (geometry validity, year/range formats, pooled/grid only, and resolution/pooling checks), surfacing invalid requests as early exceptions. 

Capability matching then uses \texttt{describe()} to verify backend (gee/local), output-mode support, and temporal semantics, and rejects mismatches with interpretable errors. For batched export, it resolves sensor configuration and uses batch inference when available, otherwise falling back to per-sample inference.

\subsection{Export}

\textbf{Output:} In its output design, rs-embed uses a unified \texttt{Embedding(
data, meta)} structure that separates numeric embeddings from contextual metadata. The data field is standardized as either a pooled vector ($D,$) or a spatial feature grid ($D,h,w$). The meta field stores reproducibility-critical details such as model identity/type, backend, sensor and temporal settings, input size, and model-specific inference parameters (e.g., pooling, bands, checkpoints, and token/grid layout). 

\textbf{Failure Isolation, Retry, Recoverability:} {\itshape Rs-embed} isolates failures at the point- and model-level. With \texttt{continue\_on\_error} on, per-item inference / sample-build / write errs will not halt the batch; results are tagged in the manifest as \texttt{ok/partial/failed}. It also supports bounded retries w/ exp. backoff (\texttt{max\_retries}, \texttt{retry\_backoff\_s}) for provider init, provider fetch, inference, and export to mitigate transient issues. For recoverability, each export unit emits a structured manifest incl. fail stage, err details, and summary stats, enabling auditable partial outputs and fast reruns for large-scale RS embedding jobs.


\section{Experiments}

\subsection{Use Case: Maize yield mapping}

\begin{figure}[h]
    \centering
    \includegraphics[width=0.45\textwidth]{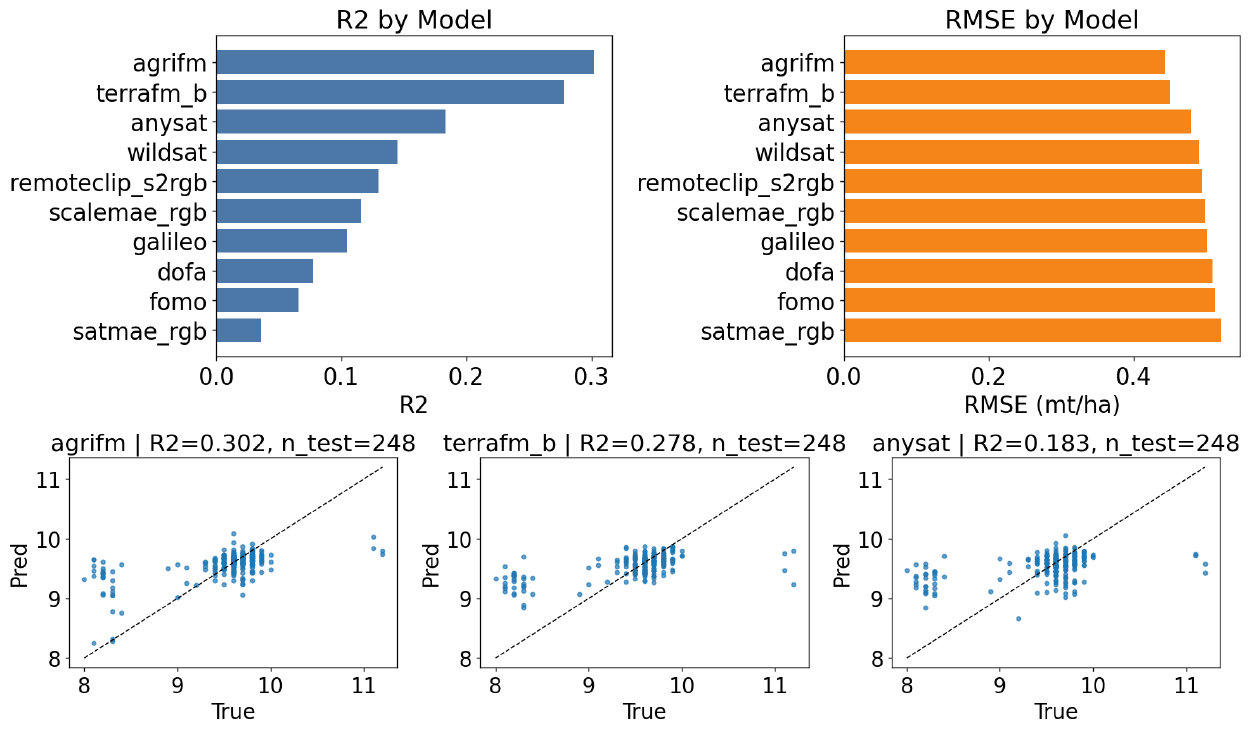}
    \caption{Performance comparison of different rsfm embeddings for maize yield regression in Illinois. The top row shows bar plots of test-set $R^2$ and RMSE (mt/ha) for each model. The bottom row presents predicted (Pred) vs. true (True) scatter plots for the top three models, with the dashed line indicating the ideal 1:1 reference.}
    \label{fig:result}
    
\end{figure}

\begin{figure}[h]
    \centering
    \includegraphics[width=0.48\textwidth]{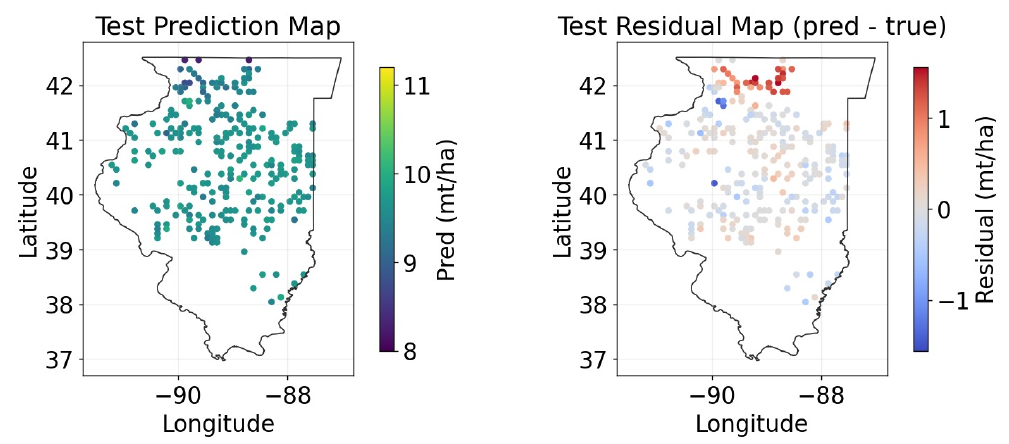}
    \caption{Spatial maps of test predictions and residuals in Illinois(Argrifm). Left: predicted maize yield (mt/ha). Right: residuals (pred - true), where red indicates overestimation and blue indicates underestimation.}
    \label{fig:residual}
    
\end{figure}

We conduct a regression experiment to predict maize yield in Illinois using embeddings from rs-embed, with SPAM2020V2\cite{DVN/SWPENT_2024} serving as the supervised label dataset. To ensure that samples are primarily drawn from cropland areas, we select regions with Maize Area > 2500 ha (more than 80\% of each field) as valid farmland samples, yielding a total of 991 sampling points. We then extract the embedding features at the corresponding locations for each model during the period from “2019-06-01” to “2019-08-01”, and train a Random Forest regressor for prediction. The results are shown in Fig. \ref{fig:result}. Overall, Agrifm achieves the highest $R^2$. However, its ability to fit samples with extremely high or low yields remains limited, and it struggles to accurately capture these outliers (see Fig. \ref{fig:residual}).

\subsection{Embedding Visualization}
\label{sec:visual}
\begin{figure}[h]
    \centering
    \includegraphics[width=0.48\textwidth]{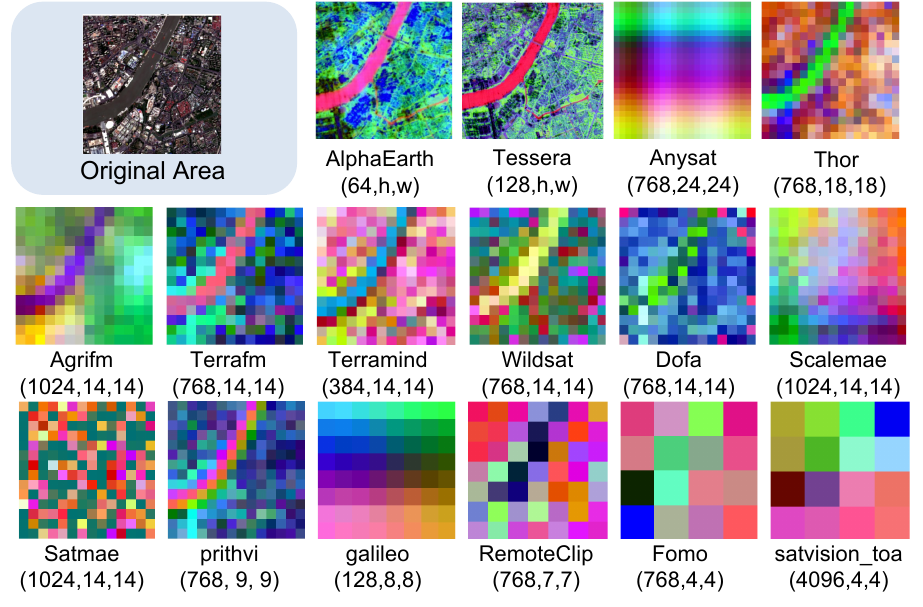}
    \caption{Visualization of embeddings from different models (\texttt{OutputSpec.grid()}, \texttt{input\_prep='resize'}). The numbers in parentheses denote the channel, height, and width. For visualization, we apply PCA and use the top three principal components as pseudo-RGB channels.}
    \label{fig:visual}
    
\end{figure}

We use rs-embed to visualize and compare the embeddings produced by rsfm under identical spatiotemporal settings. The temporal range is set to \texttt{TemporalSpec.range("2022-06-01","2022-09-01")} (for models only support yearly embeddings, we use year 2022), and the spatial location is specified as \texttt{PointBuffer(lon=121.5, lat=31.2, buffer\_m=2048)}. Under this configuration, we display the embeddings from 16 models\cite{brown2025alphaearth,feng2025tessera,forgaard2026thor,astruc2024anysat,li2025agrifm,danish2025terrafm,jakubik2025terramind,daroya2025wildsat,szwarcman2025prithvi,tseng2025galileo,bountos2025fomo,xiong2024neural,cong2022satmae,liu2024remoteclip,spradlin2024satvision,reed2023scale} using \texttt{OutputSpec.grid()}, as shown in Fig. \ref{fig:visual}.
Due to differences in training objectives and datasets, these models emphasize different aspects of spatial representation. Nevertheless, their embeddings are generally able to capture key land-cover structures to some extent, such as prominent features like rivers.

\section{Conclusion \& Future}

This paper proposes rs-embed, a unified ROI centric interface that standardizes the generation, alignment, and serving of embeddings from multiple remote-sensing foundation models. It contributes in two ways: (1) a reusable platform and benchmarking toolkit that enables consistent switching and deployment across models, reducing integration and comparison cost; (2) a spatiotemporal benchmark that compares model performance and robustness across space-time slices and provides clear feedback on how conditions, sensors, and scales affect model design choices. It also supports cross-model embedding collaboration via alignment and fusion, fostering a more open and composable ecosystem. Looking ahead, although our current focus is on remote sensing, the ROI-centric interface design are naturally extensible to broader geospatial modalities, paving the way for a unified embedding layer across sensors, data types, and spatiotemporal scales.


\begin{acks}
This research was primarily supported by I-GUIDE, an Institute funded by the National Science Foundation, under award number 2118329. This work also benefited from the Taylor Geospatial Institute RAILS system supported by the National Science Foundation under award number 2232860.
\end{acks}
\bibliographystyle{ACM-Reference-Format}
\bibliography{sample-base.bib}

\appendix
\section{Detail on use case}

\begin{figure}[h]
  \centering
  \includegraphics[width=0.5\textwidth]{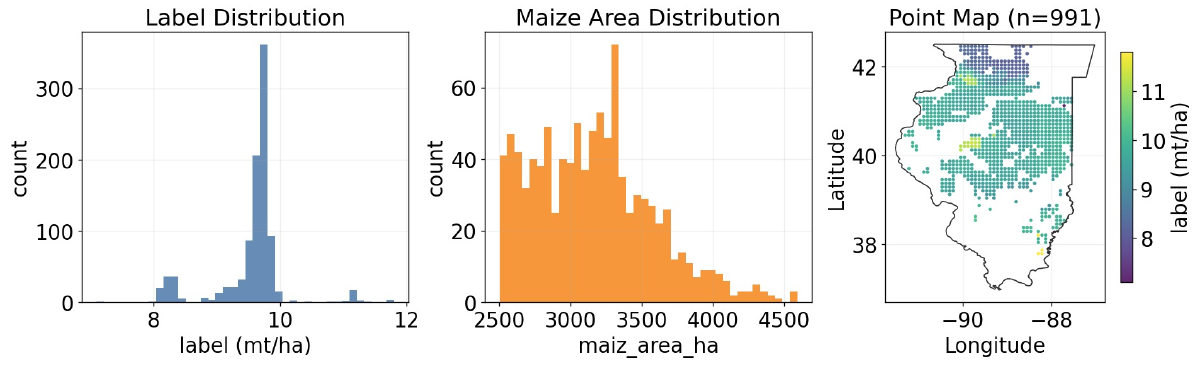}
  \caption{Dataset overview (n = 991). Left: yield label distribution (mt/ha). Middle: maize field area distribution (ha). Right: spatial distribution of sampling points colored by yield label within the study region.}
  \label{fig:sample}
  
\end{figure}

\begin{figure}[h]
  \centering
  \includegraphics[width=0.4\textwidth]{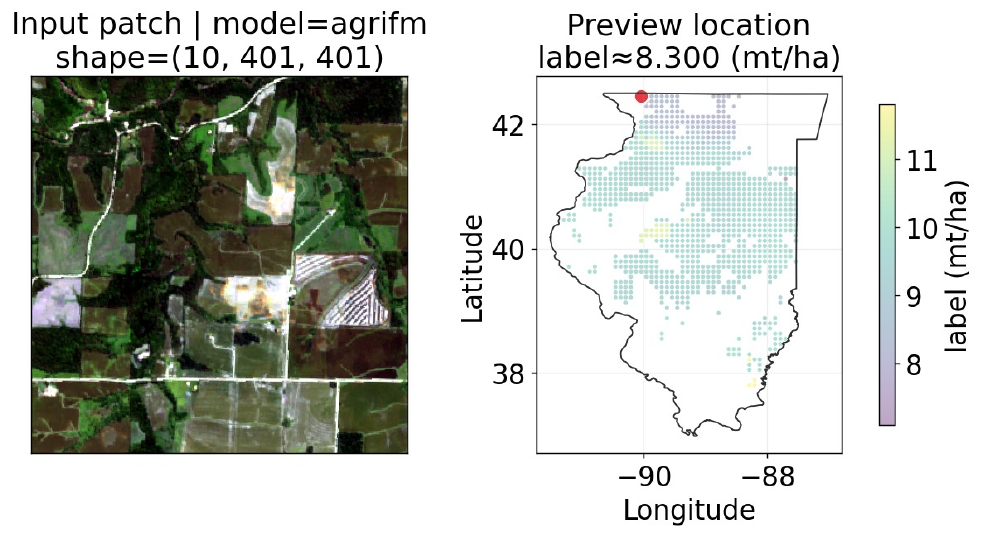}
  \caption{Sample input and spatial context.}
  \label{fig:preview}
  
\end{figure}

As shown in Fig. \ref{fig:sample}, we use a total of 991 labeled sample points. The figure summarizes the label distribution, the size (area) distribution of maize-growing regions, and the spatial distribution of sampling locations. In addition, we extract the embedding representation for each model using the pooled output from OutputSpec.pool(). Fig. \ref{fig:preview} also presents representative remote-sensing imagery for each region.

\section{Visualization on `tile` mode}
\begin{figure}[h]
  \centering
  \includegraphics[width=0.5\textwidth]{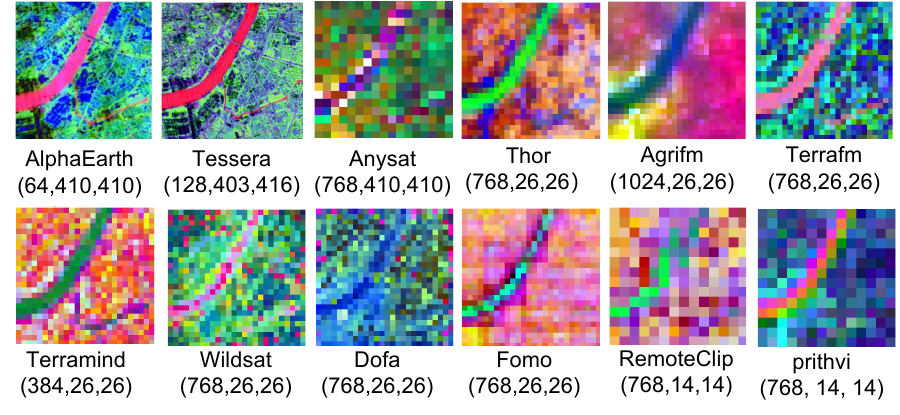}
  \caption{Visualization on tiles mode.}
  \label{fig:vistile}
 \end{figure}

`rs-embed` allows users to crop input images to the rsfm's input size based on available compute, which can produce more fine-grained embeddings. By default, inputs are `resized`. We recommend using this on models that natively support grid() outputs. Figure \ref{fig:vistile} shows a visualization of the results in `tile`' mode.

\section{Code Examples}
Below is a quick-start example for \texttt{rs-embed}. For more details, please see the \textcolor{blue}{
\href{https://cybergis.github.io/rs-embed/}{documentation}} and the
\textcolor{blue}{\href{https://github.com/cybergis/rs-embed/blob/main/examples/playground.ipynb}{playground notebook}}.

\subsection{Core API}

\begin{pycode}[title=Define Space/Time]
from rs_embed import BBox, PointBuffer, TemporalSpec, OutputSpec, get_embedding

# Spatial: point + buffer
spatial_point = PointBuffer(
    lon=121.5,
    lat=31.2,
    buffer_m=2048,
)

# Spatial: bounding box
spatial_bbox = BBox(
    minlon=121.45,
    minlat=31.15,
    maxlon=121.50,
    maxlat=31.20,
)

# Temporal: single year
temporal_year = TemporalSpec.year(2024)

# Temporal: date range
temporal_range = TemporalSpec.range(
    "2022-06-01",
    "2022-09-01",
)
\end{pycode}

\begin{pycode}[title=Get single embedding]
emb = get_embedding(
    "gse_annual",
    spatial=spatial_point,
    temporal=temporal_year,
    output=OutputSpec.grid(scale_m=10),
)
\end{pycode}

\subsection{Batch Excusion API}

\begin{pycode}[title=Get batch embedding]
points = [
    PointBuffer(lon=121.5, lat=31.2, buffer_m=100),
    PointBuffer(lon=121.6, lat=31.3, buffer_m=100),
    PointBuffer(lon=120.0, lat=30.0, buffer_m=100),
]

embeddings = get_embeddings_batch(
    "satmae_rgb",
    spatials=points,
    temporal=temporal_range,
    output=OutputSpec.grid(),
    backend="gee"
)

for i, emb in enumerate(embeddings):
    print(f"Embedding {i} shape: {emb.data.shape}")
\end{pycode}

\begin{pycode}[title=Export batch embedding]
from rs_embed import export_batch

export_batch(
    spatials=points, # multiple areas
    temporal=temporal_range,
    models=["remoteclip", "prithvi"], # multiple models
    out="exports",
    layout="per_item",
    backend="gee",
    device="auto",
    save_inputs=True,
    save_embeddings=True,
    resume=True,
    show_progress=True,
)
\end{pycode}

\end{document}